\documentclass[11pt]{article}
\usepackage{coling2020}
\usepackage{times}
\usepackage{url}
\usepackage{latexsym}
\usepackage{dirtytalk}
\usepackage{array}
\usepackage{longtable}
\usepackage[dvipsnames]{xcolor}

\colingfinalcopy %

\title{Facts2Story: Controlling Text Generation by Key Facts}

\author{Eyal Orbach \\
  Bar Ilan University / Ramat Gan, Israel \\
  {\tt eyalorbach@gmail.com} \\\And
  Yoav Goldberg \\
  Bar Ilan University / Ramat Gan, Israel \\
  Allen Institute for Artificial Intelligence \\
  {\tt yogo@cs.biu.ac.il} \\}

\date{}

\begin{document}
\maketitle
\begin{abstract}
    Recent advancements in self-attention neural network architectures have raised the bar for open-ended text generation. Yet, while current methods are capable of producing a coherent text which is several hundred words long, attaining control over the content that is being generated---as well as evaluating it---are still open questions. We propose a controlled generation task which is based on expanding a sequence of facts, expressed in natural language, into a longer narrative. We introduce human-based evaluation metrics for this task, as well as a method for deriving a large training dataset.  
    We evaluate three methods on this task, based on fine-tuning pre-trained models. We show that while auto-regressive, unidirectional Language Models such as GPT2 produce better fluency, they struggle to adhere to the requested facts. We propose a plan-and-cloze model (using fine-tuned XLNet) which produces competitive fluency while adhering to the requested content. 
\end{abstract}
\blfootnote{
    \hspace{-0.65cm}  %
    This work is licensed under a Creative Commons 
    Attribution 4.0 International License.
    License details:
    \url{http://creativecommons.org/licenses/by/4.0/}.
}

\section{Introduction}
Story generation is a challenging task in natural language processing, which requires automated systems to produce creative, open-ended text that remains coherent, cohesive and preferably engaging: an ability humans are clearly capable of.
Earlier attempts at harnessing neural techniques for story generation struggled with staying coherent for long periods of time, and proposed solutions in the form of two-stage processes which first generate skeletal structures and then expand them into longer texts \cite{harrison2017toward,event2event,planAndWrite}.
Since then, advancements in self-attention architectures and large-scale training resulted in series of pre-trained language models \cite{gpt2,grover} that demonstrate an ability to remain on topic, while generating cohesive passages which are several hundreds of words long. While the generated text is almost indistinguishable from text produced by humans, the ability to control the contents of the text remains very limited and is still an active research topic \cite{grover,ctrl,encagnostic,uber}.

Attempts to control the content of the text focus on supplying a prefix \cite{gpt2,ctrl}, a headline \cite{grover} or a writing prompt \cite{writingprompt,encagnostic}, mechanisms that can influence the overall theme of the story, but do not provide much control over its content.
Can we harness the power of large pre-trained models to generate coherent text, while allowing finer-grained controlled over the generated text?

We suggest a task in which the content of the text is specified using a sequence off \emph{key facts} expressed in natural language, which the story should follow, in order. An example of key facts \footnote{We use the term 'fact' as used in \emph{Open Information Extraction}: a factive sentence in the text, which may or may not correspond to an actual fact in the real world.} and a corresponding story appear in Figure \ref{fact-example}. This formulation suggests  natural evaluation criteria (Section \ref{evaluation-section}): the generated text should include all facts, while remaining coherent.
We show how to derive a large training corpus of fact-sets and corresponding stories (Section \ref{deriving-corpus-section}), and experiment with fine-tuning pre-trained language models on this objective (Section \ref{baseline-section}, Section \ref{proposed-approach-section}). As left-to-right generation (GPT2, BART) tend to diverge from the input and focus on the generated content (a behavior consistent with the literature \cite{writingprompt}), we propose a plan-and-cloze method, based on a fine-tuned in-place cloze model (XLNet), which performs considerably better.
Our contributions in this work are:
\begin{itemize}
\item We introduce a challenging task for controlled text generation that has a clear criteria regarding adhering to given input while maintaining high degree of freedom for the generated text.
\item We introduce a new, simple and data-efficient technique to tackle this task, by learning to produce a cloze \cite{cloze} like structure from the input, enabling the utilization of multi-directional pre-trained models, specifically XLNet \cite{xlnet}.
\item We evaluate state-of-the-art pre-trained models and techniques on this task, highlighting limitations that arise from unidirectional pre-training or difficulties to generalize to large denoising objective.
\end{itemize}

Our code and data are publicly available for future research. \footnote{https://github.com/eyal-orbach/Facts2Story-data} \footnote{https://github.com/eyal-orbach/Facts2Story-XLNetPlanCloze}

\section{Related Work}
\begin{figure*}
\begin{center}
\begin{tabular}{| m{5em}| m{13cm}|}
 \hline
 \small{Facts} & \small{
 \textbf{Fact 1:} Italian immigrant Antonio Tony Camonte kills Big Louis Costillo \newline
  \textbf{Fact 2:} Johnny takes control of South Side with Tony as key lieutenant \newline
  \textbf{Fact 3:} Tony starts ignoring orders shooting up bars belonging to O'Hara and attracting attention of police and rival gangsters \newline
  \textbf{Fact 4:} police move to arrest Tony for Guino 's murder \newline
  \textbf{Fact 5:} Cesca ends up helping him to fight police 
  }\\
 \hline
 \small{Corresponding text} & \small{In 1920s Chicago, \textcolor{red}{\textbf{Italian immigrant Antonio ``Tony'' Camonte}} acts on the orders of Italian mafioso John ``Johnny'' Lovo and \textcolor{red}{\textbf{kills ``Big'' Louis Costillo}}, the leading crime boss of the city's South Side. \textcolor{orange}{\textbf{Johnny}} then \textcolor{orange}{\textbf{takes control of the South Side with Tony as}} his \textcolor{orange}{\textbf{key lieutenant}}, selling large amounts of illegal beer to speakeasies and muscling in on bars run by rival outfits. However, Johnny repeatedly warns Tony not to mess with the Irish gangs led by O'Hara, who runs the North Side. \textcolor{purple}{\textbf{Tony}} soon \textcolor{purple}{\textbf{starts ignoring}} these \textcolor{purple}{\textbf{orders}}, \textcolor{purple}{\textbf{shooting up bars belonging to O'Hara}}, \textcolor{purple}{\textbf{and attracting the attention of the police and rival gangsters}}. Johnny realizes that Tony is out of control and has ambitions to take his position. Meanwhile, Tony pursues Johnny's girlfriend Poppy with increasing confidence. At first, she is dismissive of him but pays him more attention as his reputation rises. At one point, she visits his ``gaudy'' apartment where he shows her his view of an electric billboard advertising Cook's Tours, which features the slogan that has inspired him: ``The World is Yours.'' Tony eventually decides to declare war and take over the North Side. He sends the coin flipping Guino Rinaldo, one of his best men and also his close friend, to kill O'Hara in a florist's shop that he uses as his base. This brings heavy retaliation from the North Side gangs, now led by Gaffney and armed with Thompson submachine guns, a weapon that instantly captures Tony's dark imagination. Tony leads his own forces to destroy the North Side gangs and take over their market, even to the point of impersonating police officers to gun down several rivals in a garage. Tony also kills Gaffney as he makes a strike at a bowling alley. Johnny believes that his protege is trying to take over, and he arranges for Tony to be assassinated while driving in his car. Tony manages to escape this attack, and he and Guino kill Johnny, leaving Tony as the undisputed boss of the city. Tony's actions have provoked a public outcry, and the police are slowly closing in. Then he sees his beloved sister Francesca (``Cesca'') with Guino, and kills his friend in a jealous rage, before the couple can inform him of their secret marriage. His sister runs out distraught and tells the police what he has done. The \textcolor{magenta}{\textbf{police move to arrest Tony for Guino's murder}}, and Tony holes up in his house and prepares to shoot it out. \textcolor{brown}{\textbf{Cesca}} comes back, planning to kill him, but \textcolor{brown}{\textbf{ends up helping him to fight the police}}. Moments later, however, she is killed by a stray bullet. As the apartment fills with tear gas, Tony leaves down the stairs, and the police confront him. Tony pleads for his life, but then makes a break for it, only to be gunned down by the police. Outside, the electric billboard blazes \say{The World is Yours}.}\\
 \hline 
\end{tabular}

\caption{\label{fact-example} Training sample for the Facts-to-text task. The facts were automatically extracted from the plot of 
 \say{{\em Scarface}} (1932) using the SalIE system (Ponza et al., 2018). Note that a fact does not correspond to a contiguous span in the story. }
\end{center}
\end{figure*}
\subsection{Plotline Representations in Neural Story Generation} \label{neural-story-generation-section}
Earlier works in neural network story generation experiment with recurrent or convolutional sequence-to-sequence architectures, encountering difficulties in generating long text that stays on track. While addressing this challenge these works yield interesting mechanisms to represent the plotline of a story. ~\newcite{event2event} represent a plot as a chain of events, learning to generate a sentence from each event, while ~\newcite{planAndWrite} alternatively build the plotline as a chain of keywords. ~\newcite{checklist} avoid the need to represent each of the desired sentences by maintaining a checklist of required words and implementing a gating mechanism to insert these words and track which were used, demonstrating these abilities on cooking recipes. ~\newcite{hybrid} develop this notion further by using events as the ingredients in the checklist, but also conditioning the generated text on the desired next event, concluding that their model still generates shorter stories with less event coverage then those produced by humans.

~\newcite{writingprompt} collect a corpus of writing prompts and their appropriate stories as generated by Reddit users. Implementing a convolutional sequence-to-sequence architecture they train on generating the stories from their prompts, noting a tendency for such architectures to ignore the input and focus on the local dependencies required for language modeling rather than the more complex dependencies between the prompt and the text. To evaluate the correlation of the story to its prompt the writers suggest measuring perplexity when using the corresponding prompts versus randomly chosen ones. We claim this is not a strong enough requirement and does not suggest correlation in a semantically meaningful manner. ~\newcite{hierchalOutline} train on extractive summaries and their respective text, but do not measure the degree to which the outputs correlate to their respective summaries as they regard this technique only as a step towards improving the coherence of the generated output, concluding that human evaluations do not suggest it as helpful. 
\subsection{Text Generation with Massively Pre-Trained Transformers} \label{transformers-section}
~\newcite{gpt2} famously displayed the ability of massively pre-trained transformer models to continue prefixes of few sentences to passages of several hundreds words, that are not only grammatical, but may display thematic coherence, expanding on the same subject and adding information as the text continues. Demonstrating this capability raised questions regarding the ability to steer the content of such automatically generated text.
~\newcite{grover} have experimented with controlling the content of a news article by conditioning on metadata information such as headline or domain. Training a unidirectional auto regressive language model on a concatenated sequence of the metadata and the body, the model is capable of generating an appropriate article to a headline not seen before. ~\newcite{ctrl} train a very large scale (1.63B Parameters) unidirectional auto regressive language model on large amounts of data, prepending control codes to the text that correlate with its content or style (i.e. Books, Relationships, Legal), thus, producing a model that generates text conforming to prepended desired control codes. While displaying impressive results, both approaches requires a very large amount of resources, and are limited by either the loosely defined relation between metadata such as headline and the content, or the finite possible combinations of existing control codes.

More recent works \cite{uber,encagnostic} have experimented in how to utilize an already pre-trained unidirectional model like GPT2 \cite{gpt2}. Their approach argues that if this model can self-attend to the preceding content in a way that controls its next output, then it is possible to produce the representations the model attends to, without the need for explicit text. ~\newcite{encagnostic} demonstrate the ability to encode different problem-domains to representations that can be appended to the context the model attends to, in order to produce control of the desired text. The large flexibility of this approach in regards to the supplied input makes it a viable candidate for our task while our experiments highlight some of its limitations.
Both ~\newcite{bart} and ~\newcite{t5} have taken an approach of shaping the unsupervised objective of language models to be a sequence-to-sequence denoising objective, where the input is some corruption of the text and the expected output is the original text. These works which demonstrated high performance on several text generation benchmarks, are also suitable for our proposed task and so we experiment with BART \cite{bart} in our desired setting. 

\section{Stories and Facts}
In contrast to the earlier works in neural story generation referenced in section \ref{neural-story-generation-section}, our main focus in this work is the degree to which generated stories can adhere to the desired content, as we harness the power of massively pre-trained self-attention models, to mitigate the difficulty these works encountered with keeping long texts on track. Additionally we differ from previous works that explore controlling massively trained transformers (section \ref{transformers-section}) in that we focus solely on fine-tuning, and represent the desired content in a more explicit, semantic meaningful manner, that is easy to use, modify and evaluate, and encompasses content that should appear throughout the text as opposed to various forms of seeding.

We therefore seek a method for control-over-content, with a representation of the desired plotline that satisfies both the following requirements:
\begin{itemize}
    \item The representation should be minimal enough to require the system to substantially expand on it in order to create a full story, remaining in the territory of open-ended generation as opposed to more confined text generation settings like translation or data-to-text tasks.
    \item The representation should be explicit enough in its nature to meaningfully affect the entire generated story and to effectively determine if the content has sufficiently conformed to it.
\end{itemize}

 Event representations that have been used in previous works, are considerably structured, unnatural for non experts, verb-centric, and are very minimal in information, adding challenges to the transition to natural language that are unnecessary for our setting. On the other hand, natural sentences vary significantly in length, may fuse a large number of events, employ various discourse markers and include other complex phenomenas which we would like to minimize. We suggest the use of an intuitive, loosely defined notion of \emph{key facts}, simple sentences that describe factual information in the world of the story. Each fact should describe an event that occur in the narrative (\emph{Dorothy's house is swept away by a tornado}), state the characteristics of a person or place (\emph{The magical forest is haunted by gruesome trolls}), specify an emotion experienced by a character (\emph{The queen is jealous of Snow White}) and so forth. Consequently, this more concise nature of \emph{facts} provides a larger degree of freedom in the mapping from facts to complete sentences in a surrounding context.  \emph{Key facts} accordingly, are primary pieces of information that are central to the full desired plot (see figure \ref{fact-example}).

\section{The Task}
Given an input of 5 ordered \emph{key facts}, we wish to generate a story 100 to 1000 words long, that includes all the facts given as input in their order of appearance, while expanding upon them to produce a broader, yet coherent, narrative. 

\subsection{Evaluation Metrics} \label{evaluation-section}
Our setting of creative story generation inherently lacks a \say{truth} text to compare to aside from the input facts, limiting the ability to use metrics like ROUGE \cite{rouge}. Further we refrain from using perplexity to compare between the models since a custom language model trained only on our corpus will significantly underperform our utilized pre-trained models, and an LM pre-trained on a sufficiently large corpus, introduces hard to control bias, due to the different pre-training data of the compared models.

Instead, we perform human-based evaluation by Amazon Mechanical Turk workers, based on three evaluation criteria, described below.
\begin{itemize}
    \item \textbf{Adherence to Facts} \newline
    The worker is supplied all 5 input facts and is required to state for each fact if it can be found in the generated text, either verbatim or rephrased.
    \item \textbf{Grammatical Correctness} \newline
    Supplied with examples and a qualification test on what is grammatical and what is not, worker is asked to rate the grammatical correctness of the text on a scale of 1 to 5.
    \item \textbf{Common Sense} \newline
    The worker is asked to rate the plausibility of events described in the text in regards to the context and the events preceding them. While being elusive to define, it is common to expect sufficiently high human agreement on the plausibility of events to follow one another \cite{mcrae2005semantic,ordinalCommonSense}. We note here that the knowledge required to determine if a chain of events is plausible in a fictional story requires not only understanding of the \say{real} world, but also of what constitutes narratives  and the complex relation between stories and reality. We supply the worker the example in Figure \ref{fcommonsense-example} along with a short explanation to convey what the worker is expected to rate in this criteria.
\end{itemize}

\begin{figure*}
\begin{center}
\begin{tabular}{| l| l|}
 \hline
 \textbf{Plausible} & 
 \emph{After dying Tony comes back to life to haunt Kate.}\\
 \hline
 \textbf{Less plausible}  & 
\emph{After dying Tony comes late to his football practice.}\\
 \hline
\end{tabular}
\end{center}
\caption{\label{fcommonsense-example} Examples of ordered events with different plausibility in the context of a movie plot.}
\end{figure*}

\section{Deriving Appropriate Corpora} \label{deriving-corpus-section}

Traditional approaches to narratology often separate between the story as a sequence of events and  techniques of telling the story \cite{culler_2001}. In this work we focus on descriptions of movie plots as they express in natural language clear series of events, providing stories with a large variance in subject matter, yet that conform to a similar narration approach, as opposed to the large variety found in literature works that employ more complex techniques like dialogues, first person story teller and so forth. 
We use a large corpus of movie plots taken from Wikipedia \cite{movieplotsDataset,filmCharchters} as they also conform to similar lengths and format. 

\subsection{Test set} Since the pre-trained models we evaluate in this work, have been exposed to a large portion of Wikipedia, and these models and weights were all published before June 2019, we collect a test set of 100 movies published after that date, as the models were not exposed to these plotlines at pre-training. We manually extract key facts for each of these plots, and regard these fact-sets as test inputs, with the original stories now serving only as a reference (which will not be used in evaluation).

\subsection{Training set}
In addition, we wish to derive a much larger dataset containing thousands of examples, to enable supervised fine-tuning. While manual facts extraction is not feasible, the facts format correlates nicely with the artifacts of OpenIE frameworks \cite{BankoOpennIE} and we therefore employ existing frameworks \cite{minie} to automatically extract facts from the large Wikipedia-based plots corpus. As we desire a limited number of only 5 key facts per story, we take advantage of the SalIE framework \cite{factsThatMatter} that aims to rate saliency for extracted facts, and use the derived facts with the highest saliency scores as the key facts.
Using these frameworks, described in \emph{Technical Details} below, we derive a training data set of 17 thousand sets of key-facts and their correlating stories. The derived corpus has a ratio of 1/6 between the number of words in the key facts and the full plot, requiring the model to produce substantially more text then the context it is exposed to: the unsupervised denoising objectives used in pre-training the models addressed in this paper, as well as other notable models, typically train while masking only between 15\% to 30\% of the tokens \cite{xlnet,bart,bert} providing substantially larger visible context. We also note that our training data is noisy, as these automatic frameworks make various mistakes with either the roles or boundaries of a fact's constituents or the attempt to rebuild the fact as a proper sentence. Therefore the training data includes key facts that are not always grammatical, yet are still useful enough to facilitate supervised training.

\paragraph{Technical Details}
\emph{Open Information Extraction} works commonly incorporate the notion of facts with a structured, more strict definition, as tuples of the form $\langle$Entity1, Relation, Entity2$\rangle$, suggesting systems that optimize the objective of extracting all such facts from heterogeneous domains. ~\newcite{clauseIE} developed ClauseIE, an \emph{Open Information Extraction} framework that leverages knowledge about English grammar to deal with the complex structures of sentences, finding multiple clause candidates in an input sentence and generating relevant synthetic clauses by recognizing relative pronouns, possessives and other known phenomenas. MinIE ~\cite{minie} is an improvement built on top of ClauseIE addressing the drawback of proposed constituents being overly specific by introducing semantic annotations like quantity and polarity. We use the extracted tuples of MinIE---converted back to a sentence form---as our desired facts.

\emph{Saliency:} ~\newcite{factsThatMatter} built upon these frameworks defining an objective to evaluate the facts' salience scores, determining how essential is its information to the message the text conveys. Their framework, SalIE, rates facts extracted by MinIE and aims to output the most salient ones while maintaining sufficient diversity.
These outputs alone, are shown to perform competitively on text summaries metrics like ROUGE. To infer a fact's relevance, the framework implements a variation on the idea of the PageRank algorithm ~\cite{pagerank}, where facts are vertices for which averaged Glove embedding is calculated. Cosine similarity between vertices specifies the weight of a corresponding edge and the facts position in the text serves as a relevance prior. 
We use the top-5 salient MinIE extractions, as rated by the SalIE framework, using the framework's mechanism to present these extractions in the form of a sentence by concatenating subject, relation and object strings and returning necessary words removed with MinIE's annotations, as seen in figure \ref{fact-example}.

\section{Baselines: Sequence-To-Sequence Models} \label{baseline-section}
The facts-to-story task requires conditioning on the input facts. This is naturally achieved in sequence-to-sequence models like BART, as well as prefix-conditioned models like GPT-2. 
Our baselines are thus using these models while fine-tuning them on our training set.

\paragraph{Fine-tuned GPT2 + Pseudo Self Attention}
While GPT2 \cite{gpt2} has demonstrated impressive generation abilities, its unidirectional left-to-right paradigm, introduces several challenges. Our preliminary experiments with concatenating the salient facts as a prefix for the generated text, has failed to converge to reasonable results on our training data.
Attributing this to vast difference in structure between the key facts and the consecutive prefixes the model has been exposed to during pre-training, we resort to the technique of \emph{pseudo self attention} as proposed by \newcite{encagnostic}. We follow their proposed architecture for story generation on our custom input, hence learning an encoder that can map the key facts input text, to representations that are concatenated to the GPT2 decoder hidden states, therefore enabling the decoder to attend to the input while learning to map it effectively. The input for this encoder is the 5 key facts in sentence form, concatenated consecutively, where in addition to the token embeddings and the position embeddings, we add fact embeddings, that distinguish between the 5 different facts and are learned as part of the fine-tuning process. These facts are added to every token within each fact similarly to the sentence embeddings of \newcite{bert}.

\paragraph{Fine-tuned BART}
BART \cite{bart} is a sequence-to-sequence model similar in architecture to the original Transformer \cite{attentionisallyouneed}, pre-trained on a de-noising objective where the input to the encoder is a corrupted version of the expected output. There are several corruption mechanisms that were applied to the input text during pre-training, where the two mechanisms most relevant to our setting are the deletion of random tokens as well as the replacement of random tokens and spans with a custom [MASK] token. This de-noising paradigm fits rather naturally to our proposed task and so we supply as input the concatenated facts in their sentence form, separating between the different facts with the [MASK] token. This setting requires the model to generalize phenomenas it was exposed to in pre-training to a larger number of missing tokens, while also appropriating the [MASK] token as a signal to separate between the different facts.

\section{Proposed Approach: Plan and Cloze} \label{proposed-approach-section}
Both of the above suggested approaches, essentially consist of a unidirectional, left-to-right, auto-regressive language model, conditioning itself both on all its previous outputs as well as the supplied input. Our experiments (Section \ref{Experiments-section}) suggest a limited ability of this approach to conform sufficiently to the conditions supplied in the input. Therefore, we diverge from the seq-to-seq approach and re-frame the problem as literal text expansion. Our approach consists of a trained planning step, followed by a ``fill-in-the-blanks'' cloze step.
The planning step builds a cloze-structure template, placing all tokens supplied in the input at specific positions in the output. This determines both the incorporation of fact tokens into more complex sentences, as well as the spacing between facts.
Having a hard constraint on the existence of these tokens in their selected positions, the second step forms a sensible story by filling in the blanks in the contrived template. This target of filling in blanks with hard constraints on existing words in specific positions, correlates well to the de-noising objectives of \emph{Masked Language Models} \cite{bert,spanbert} or a \emph{Permutated Language Model} like XLNet \cite{xlnet} that receive the exact structure of the target text. We select XLNet, as it is also  auto-regressive on a permutated order of predictions, aligning to the common decoding strategy of text generation.

\begin{figure*}
\centering
\begin{tabular}{|m{5em}| m{13cm}|}
\hline
\small{Facts} & \small{\textbf{Fact 1:} \emph{Farmer John and dog Bingo discover landing of UFO} \newline
\textbf{Fact 2:} \emph{Shaun discovers a trail of pizza crusts and encounters the alien} \newline [...]}\\
\hline
\small{Concatenated (Tokenized) Sequence} & \small{\textcolor{red}{\textbf{'\_Farm' 'er' '\_John', '\_and', '\_dog', '\_Bingo', '\_discover', '\_landing', '\_of', '\_UFO'}}, \textcolor{orange}{\textbf{'\_Shaun', '\_Discovers', '\_a', '\_trail', '\_of', '\_pizza', '\_crust', 's', '\_and', '\_encounters', '\_the', '\_alien'}}, [...]} \\
\hline
\small{Position predictions} & 4,0,0,0,0,3,0,0,0,36,0,0,0,0,0,0,0,0,0,0,32, [...] \\
\hline
\small{Cloze \newline Structure} & \small{\_ \_ \_ \_ Farmer John and dog Bingo \_ \_ \_ discover landing of UFO \_ \_ \_ \_ \_ \_ \_ \_ \_ \_ \_ \_ \_ \_ \_ \_ \_ \_ \_ \_ \_ \_ \_ \_ \_ \_ Shaun discovers a trail of pizza crusts and encounters the alien \_ \_ \_  [...]}\\
\hline
\small{Generated Text} & \small{In the town, \textbf{Farmer John and dog Bingo} get lost and \textbf{discover landing of UFO}. Shaun and Bingo take shelter in ``The Bake'' a large, open-air flat that contains a series of shops. There, they find a child and father. \textbf{Shaun discovers a trail of pizza crusts and encounters the alien}, Bitzer. [...]}\\
\hline
\end{tabular}
\caption{\label{facts2text} From facts to generated text (full text in appendix).}
\end{figure*}

\paragraph{Planning stage.}
We wish to learn to construct a cloze structure given only the text of the key facts. Since our training data is comprised of facts extracted from the original text, the words in the fact correlate to specific positions in the text. It is important to note that due to the clause rich nature of natural text, facts may appear in a non consecutive form, having other clauses or words appear in between the words that constitute a chosen fact. To support this we predict the \emph{spacing between the words} of each fact along with the \emph{spacing between the different facts}. To effectively learn this spacing we assume the following:
\begin{itemize}
    \item Ordering supplied in the input is reliable both referring to words inside the fact as well as between the facts.
    \item Words' meaning is viable information for appropriate spacing.
    \item Differentiating between the facts is significant for spacing prediction.
\end{itemize}
Under these assumptions, we define a learning task that focuses on a desired spacing for each word (or more precisely, for each BPE token). Given one long sequence of all the facts concatenated, each token $t_i$ of this text is mapped to a number signifying the number of tokens in the original text between it and $t_{i-1}$ (see Figure \ref{facts2text}). We capitalize on the pre-trained XLNet model's knowledge of words and context, to map a sequence of words to a sequence of positive numbers. The input to the model is the concatenated sequence of all facts with the positional embeddings sequential from 1 to $|factsTokens|$. To assist the model to differentiate between the facts  we add 5 ``fact embeddings'' that are randomly initialized and are learned as part of this task, each appropriate fact embedding vector is added to all encoded tokens that are part of the corresponding fact similar to the sentence embeddings described in \cite{bert}.  We replace the last layer of the model with a new layer of $embedding\_vector\_size \times 1$ and apply a ReLU activation function to obtain non-negative outputs. The corresponding label for each token is the actual spacing in our dataset between this token and its predecessor. We use a modified MSE loss, that penalizes mistakes on small numbers more than mistakes on large numbers: 
 \[\mathcal{L}(y,y')=\frac{(y-y')^2}{log(y)}\]
 This follows the observation that 
 adjacent words (i.e. \{should, not\}) split far apart by mistakenly predicting a large number have a more severe impact than a mistake on words that were originally located far apart. In inference time we round each output down to the closest non-negative integer.

\paragraph{CloZE-based generation stage.}
Filling in missing tokens in a cloze like structure correlates to the original pre-training objective of XLNet with the exception of the much larger number of missing tokens. %
To account for this difference from its training condition we fine-tune XLNet, forgoing the random permutation order, and instead committing to a left-to-right order for all missing tokens. Formally, given a set of all indices representing the tokens of the extracted facts denoted as $s$ our training objective is to maximize the likelihood for:
    \[\prod^{T}_{i=1, i \notin s} p(x_i|x_{<i},x_{\in s};\theta) \]
    
Combining the above steps results in a passage that contains the tokens of the original facts surrounded by generated text as can be seen in Figure \ref{facts2text}. 

\section{Experiments} \label{Experiments-section}

\subsection{Models Training and Inference}
We hold out one ninth of the training material to serve as a validation dataset to avoid overfitting during fine-tuning.
Inference decoding strategies and their parameters vary, and were set individually by starting from the package's default values and tweaking parameters while manually observing the results

\paragraph{GPT2 + Pseudo Self Attention.}
We use the code published by \newcite{encagnostic} utilizing its support for additional embeddings and train with the configuration for story generation, with the 117M parameter GPT2 model and additional parameters for the encoder resulting in 181M parameters. We train until minimal perplexity is reached on the held-out validation set. Generation for the test evaluations is done with top k sampling of 100 and temperature of 0.9.
\\\noindent\textbf{BART.}\hspace{0.7em}
We fine-tune BART-Large with the implementation available in the Transformers framework \cite{transformers} with its 406M parameters, fine-tuning until reaching minimal loss on the validation set. Generation for the test evaluations is done with nucleus sampling \cite{nuculus} value of 0.9 and 2.5 repetition penalty.
\\\noindent\textbf{XLNet with Cloze Structure Prediction.}\hspace{0.7em}
For text generation we use XLNet-base with its 117M parameters and for the position prediction we use a separate XLNet-base model with an additional self-attention layer resulting in 123M parameters. Generation for the test evaluations is done with top-k sampling of 40 and temperature of 0.85.

\subsection{Evaluation}
For evaluation, we apply the three models to 50 test-set items (resulting in 150 generated stories), and evaluate each generated text for the criteria presented in Section \ref{evaluation-section} using Amazon Mechanical Turk.
To mitigate the issue where different raters may be inclined towards higher or lower ratings, a rater's task was to rate all 3 stories generated from a fact-set (one story from each system), presented in random order. On top of this,
each 'facts-to-3-stories' set was evaluated three times by three different raters. This method balances out scaling issues between raters, and allows us to regard the average rating as a reliable evaluation metric. In total 58 different anonymous evaluators participated, which have passed a simple grammatical correctness qualification test.

\begin{table*}
\centering
\begin{tabular} {||l | c c c||} 
 \hline
 Model & Grammar & Common Sense & \# Facts Found \\  
 \hline\hline
 Pseudo Self Attention GPT2 & \textbf{3.77} & \textbf{3.31} & 0.75 \\ 
 \hline
 BART & 2.68 & 2.58 & 1.61 \\
 \hline
 XLNet Plan + Cloze & 3.48 & 3.25 & \textbf{4.39} \\ 
 \hline
\end{tabular}
\caption{Averaged rating per model\label{results}}
\end{table*}

\subsection{Results}
 \paragraph{Quantitative results.} The results in Table \ref{results} show that our proposed plan+cloze solution produces the largest number of facts by a significant margin, while being competitive in regards to grammar and common sense. 
 Interestingly, despite its pre-training objective, BART incorporates only marginally more facts than GPT2, while performing substantially worse than both models on both grammaticality and common-sense.
\paragraph{Qualitative Analysis and Discussion.} \label{analysys-section}
Some generated examples for each model appear in the appendix. We manually inspected a large sample fact-sets and their generated stories.
 It is interesting to note that the quality of the generated text proved highest for GPT2, and we attribute that in large to its left-to-right unidirectionality. Analyzing the text produced, exposes that the pseudo-self-attention mechanism was successful in leading the decoder towards production of entity names that appeared in the input facts, along with various themes, like \say{aliens} or \say{murder} but failed to either copy or represent in other ways, relations described in the facts, producing text that is more loosely related to the input than intended.
 We claim this can also be attributed to the left-to-right unidirectional pre-training, requiring substantially more \say{retraining} (in terms of time as well as data) for the model to learn to attend to representations that signify text that is in the \say{future} or to the right of the text being generated. 
 
 While BART has been designed to leverage the advantages of unidirectionality in its decoder and multidirectionality in its encoder, it displayed the worst performance in contrast to its reported high relative performance in various generation tasks. While the reasons for it require further investigation we note that the 2 major differences in our setting are the requirement for relatively long generated text along with smaller amount of exposed text in relation to the text to predict. \newcite{t5} report degradation in performance on several tasks on a similar architecture to BART when raising the corruption rate to 50\% which is still smaller than our setting which is 84\% on average. Analysing its outputs exposes exceptionally long sentences. The first fact usually appears in the generated text verbatim, while elements of the second fact may appear oftentimes losing their original meaning. All other facts seem to have very little effect if any on the generated text, suggesting the dependencies of the decoder are easier for the model to pick up and rely on, than the more complex, forward looking dependencies of the text supplied to the input. This phenomena of sequence-to-sequence models focusing primarily on the dependencies of the decoder, in settings where the supplied input is substantially smaller then expected output has also been observed in previous works \cite{writingprompt}. 
 
XLNet pre-training involved predicting only 15\% of the tokens while being exposed to the rest, which is twice less than BART's pre-training objective, yet while also having a smaller parameter count, our suggested cloze-XLNet method surpasses BART on all our reported metrics. Our position prediction approach enables to explicitly state the facts while creating constraints that help the generation process remain on track.
Due to substantial training in such setting, XLNet is apt to adhere to local constraints making the text transition smoothly between the copied facts and the generated text. When a single fact's tokens are spread out, the original meaning usually remains, with the added tokens expanding the sentence in a sensible way. When the constraints of the various facts' tokens enable a very large degree of freedom for the generation process, we notice several recurring mistakes in XLNet's generated text. It is common for entities to switch gender throughout the text. Entities may also tend to switch roles appearing incorrectly both as the subject and the object of a relation (\say{The alien, Bitzer, is a drug smuggler who has a plan to make Bitzer a human}). It is also apparent that the order of the events depicted is rather arbitrary, excluding the opening and ending of the text that conform to repeating patterns. Further research is required to determine if unidirectional generation in training helps mitigate these issues and to what degree.

\section{Conclusion}
We suggest a challenging task of controlling story generation content by \emph{key facts}. We demonstrate an approach for deriving a large corpus for this task, suitable for supervised fine-tuning, and evaluate the ability of notable massively pre-trained models, fine-tuned on this corpus, to generate stories adhering to our suggested representations of content. We show a simple planning technique of copying and structuring, that enables reframing the objective as formulating a cloze task followed by filling in the blanks with a pre-trained, permutated order, autoregressive language model, thus achieving competitive results in regard to the coherence of the generated text and substantially superior performance in adhering to the \emph{key facts} controlling mechanism. 

\bibliographystyle{coling}
\bibliography{coling2020}

\newpage
\section*{Appendix - Generation Examples}
\centering
\begin{longtable}{|m{5em}| m{13cm}|}
\caption{Generated Stories for Facts Set 1}\\  
\hline
Facts & \textbf{Fact 1:} \emph{Farmer John and dog Bingo discover landing of UFO} \newline
\textbf{Fact 2:} \emph{Shaun discovers a trail of pizza crusts and encounters the alien} \newline 
\textbf{Fact 3:} \emph{The visitor introduces herself as Lu-La, an impish alien from To-Pa with extraterrestrial powers} \newline 
\textbf{Fact 4:} \emph{Lu-La transmits her memories to Shaun} \newline 
\textbf{Fact 5:} \emph{Shaun and Lu-La restart the ship and manage to escape with Bitzer and head to To-Pa}\\
\hline
Cloze XLNet & In the town, \textbf{Farmer John and dog Bingo} get lost and \textbf{discover landing of UFO}. Shaun and Bingo take shelter in ``The Bake'' a large, open-air flat that contains a series of shops. There, they find a child and father. \textbf{Shaun discovers a trail of pizza crusts and encounters the alien}, Bitzer.  After he takes his daughter, Lu-La, from home, he finds a woman, Lu- La. He is attracted to her, but she is terrified when Bitzer arrives and asks for help. \textbf{The visitor introduces herself as Lu-La, an impish alien from To-Pa with extraterrestrial powers}. The alien, Bitzer, is a drug smuggler who has a plan to make Bitzer a human. He uses his abilities to evade and carry out the same mission. After Bitzer falls in love with the young woman, and passes out on her, \textbf{Lu-La transmits her memories to Shaun Shaun and Lu-La restart the ship and both manage to escape with Bitzer and head to To-Pa.} They join the Air Force. In the Air, they find a doctor and the elderly, who are treated by a ``Skin''. A doctor, however, is also awaiting them. A ``T-Tapper'' (Man-O), however, is waiting for them. The T-Tapper is a Nazi whom the British have captured. After they locate them, they return to the air, where they are met by a military (British) operative. They are captured and sent to the RAF. There, they are reunited with the young woman and the child. The elderly, who had been smitten, returns to the Army and the young woman. One night, they are attacked by a group of thug-like ``T-Tapper'' who go off as invaders. They kill them, but are trapped in a cave. Shaun and the young woman are rescued.\\
\hline
BART &  Farmer John and his dog Bingo discover landing of UFO. The aliens appear to have been monitoring Earth's weather patterns, but their presence leaves the locals suspicious – except for pizza-loving Shaun Kringle (Zach Galifianakis).Shaun discovers a trail that leads him back over 3 hours ago on Pizza Planet where he encountered another humanoid alien who claims her ship was hijacked by greedy bankers after she received payment from them in exchange foiling an investment scam involving one Jack Pearson; they are now convinced both parties were responsible when it crashes into nearby buildings causing heavy damage as well with all life aboard being lost along side its occupants: A homeless man named Bitzer is blamed while several police officers lose control during searches searching For Pete Elliott at shopping malls until finally killing LuLu before escaping via balloon through city streets under street lights due excessive demand caused many mall shoppers suffering mass panic attacks because food prices increased dramatically overnight following reports about illegal imports arriving nationwide. While returning home two other people encounter what appears be someone else living atop Toppan City Building outside Phoenix\\
\hline
GPT2 with pseudo-self-attention & It is the story of an imaginary cat named Luvie who disappears from her rightful residence in the Qin Shiyanan system after finding her lost husband, Dr. Steven Wu. After finding her missing husband, Luvie decides to find her husband and bring him back to her native Shaan. She meets the mysterious Zengong Luvie and becomes fascinated. She eventually recognizes her husband as the mysterious ``Big Zengong'', a ``lost comet'', which lands on Earth in the near future. After encountering the aliens, she returns to her home in an abandoned NASA spacecraft, where she meets her ex-husband and brother Alex, and his older sister, Lola. Luvie encounters Zengong Luvie, who is still trying to find her, and learns that her husband has been missing for years, and that he is actually the alien God, the ``Big Zengong.'' Lola and Luvie set out to find the missing man, after she discovers Zengong Luvie performing a dance with her former lover, Liu Wan. Luvie and Luvie eventually corner the aliens, who proceed to attack Lola, his sister, and Alex. The aliens arrive at the abandoned spaceship, but Luvie's family members, who are also aboard, fight and kill the aliens. Lola and Luvie reunite with Liu Wan, who is still trying to find Liu. The aliens soon move into a strange house, and Luvie realizes that she has been tricked. A young woman named Rebecca, who is the granddaughter of a Nazi scientist, visits her home, offering a place to stay at her own house, but Luvie refuses. Rebecca attempts to convince the aliens to let her stay, but is rejected, instead being disenchanted by the current inhabitants who use her as a hostage. The two learn that the spaceship belongs to a mysterious woman named Zengong Xuan, and the only way to get out is to enter the ship. Luvie explains to Rebecca that she was once the only one who could find her husband, allowing her to return to her rightful residence in Shaan. With Rebecca now part of the group, Luvie and Luvie go to Zengong Luvie's house. As he tries to enter, Zengong Luvie's family members, who have found the ship, are attacked by a mysterious force that uses Luvie to possess its power. The aliens are forced to enter a wormhole to escape, but Luvie manages to bring the aliens to the wormhole. However, Rebecca and Rebecca's family members are unable to escape, so Luvie and Luvie fight the wormhole. As Zengong Luvie breaks into his house and enters the wormhole, he is attacked by a giant squidtailed creature, who is then killed by the Big Zengong. As Luvie and Luvie get out, Rebecca gives birth to a baby girl named Mango. Luvie returns to her place and starts seeing Rebecca once again. Rebecca soon discovers that Luvie has transformed from a young woman into the beautiful Zengongian woman after he defeated her father. As the baby girl grows, she realizes that it was the Big Zengong; the ``Grand Zengong'' is actually a powerful force living in outer space. Rebecca, who has discovered the truth, seeks help from the Great Zengong, who offers her the help of Zengong Luvie's houseguest, Zhi Lei. As she gives them the information that Luvie is the Grand Zengong, the Great Zengong gives her a special gift that tells her that if she doesn't come back into the reality, she, too, will no longer be sent out of the universe. In time, Luvie is able to use her powers to save Rebecca and save Mango, as well as save the baby girl. Meanwhile, Rebecca is abducted by Zhi Lei, and Luvie escapes with Rebecca. Although the alien God has regained the opportunity to control one of the worlds, he is still working on the other. When Rebecca comes back into the future with Luvie, the alien God's plan is foiled by his own plan to conquer all of the universes. The two reunite with Rebecca, who becomes a mother figure to her grandchildren. Although she marries Rebecca, Luvie continues to live in the past, but remains deeply traumatized by the events that took place. In the present, Rebecca is seen trying to tell Luvie that the aliens are coming out of the universe. Luvie and Rebecca begin to cry, and Rebecca explains to Luvie that the aliens will be coming out again in time. \\
\hline
\end{longtable}

\newpage
\centering
\begin{longtable}{|m{5em}| m{13cm}|}
\caption{Generated Stories for Facts Set 2}\\  
\hline
Facts & \textbf{Fact 1:} \emph{German army has pulled from sector of Western Front in northern France} \newline
\textbf{Fact 2:} \emph{Blake cross no mans land to reach the abandoned German trenches} \newline 
\textbf{Fact 3:} \emph{German plane shot down in flames} \newline 
\textbf{Fact 4:} \emph{pilot is shot dead by Schofield} \newline 
\textbf{Fact 5:} \emph{Schofield chooses to part with them at bridge}\\
\hline
Cloze XLNet & In July 1930, on the eve of World War I, a train carrying German prisoners belonging to the \textbf{German army}, \textbf{has pulled from sector of Western Front in northern France}. Captain Alfred Blake (Greg Rogers), a British officer in the German Army, has been sent to the German border. After being briefed by one of the German troops, \textbf{Blake cross no mans land to reach the abandoned German trenches}. He is captured, but finds the \textbf{German plane shot down in flames}. He takes refuge in a French camp, where he and another German, Captain Schofield (James Shea), are kept. Blake is not informed that he is the agent. Blake is able to convince Schofield to take him to the German border. The \textbf{pilot is shot dead by Schofield Schofield chooses to part with them at bridge}, and is able to get the Germans to open a mine. After crossing the border, he finds a German in one of the German trenches, who is dead and in grave condition. He returns to the base and discovers that Schofield has become a spy. When he discovers that the German agents have been killed, he leaves Schofield in hiding. However, the man is the agent, and tells him that the train is going to the Germans. Blake is able to kill the agent by radio, and escape. As he is about to return, Schofield tells him that he has a plan to steal the railway. However, von Schofield has already stolen a train that comes to the Germans. When a German officer, Major Schlieiber (John F. Henner), arrives in the town, he is recognized by von von Schofield as a Nazi operative, and he is sent to the front to fight.\\
\hline
BART &  In 1916, the German army has pulled from sector of Western Front in northern France. Private Blake (John Mills), an English officer on leave at a British camp is reunited with his friend and fellow trainee Major Schofield(Barry Harris). While they attempt to find some food for their weary minds back home after weeks away by sea crossing into occupied Belgium via Dunkirk as well several hours walking along deserted country roads that cross no mans land trying desperately but futilely attempts made through snow drifts either side - all without success despite blithe efforts during morning light or twilight search parties flanking them everywhere -- two Germans are killed while attempting unsuccessfully another across this barren stretch; Captain Von Friese's car stolen overnight before it can be retrieved when he tries one last time earlier than planned just north-west over Normandy – whilst Lt Tom Tannenby must try yet again twice more plus retracing three days' walk each way alone up ditches not only where soldiers have fallen ill following repeated failure thus far due injury sustained under duress both physically \& emotionally since leaving England many years previously) They successfully enter French lines behind enemy guns near La Boheme forest guarded daybreak upon spotting Allied aircraft silhouetted against sunlit hills overlooking Paris even though dusk falls soon afterwards.\\
\hline
GPT2 with pseudo-self-attention & In 1940, German aircraft is shot down in a mountainous French-held forest by two German aircraft. The survivors, including a scientist named St. Clair (Michael Caine), get separated, as is the German pilot's parachute. Major St. Clair (Gary Cooper) leads his British troops ashore to the British ruins of the Rhine River which is to be destroyed by Allied forces. In late 1943, German captain Reinfeld Schlosser (Michael Moore), a leading American correspondent, takes a job on the Canadian front. The Americans have also been ordered to get stranded, so the Germans are unable to land. A few days later, St. Clair and his wife Joyce (Jane Wyatt) are rescued from their German camp and brought back to the Allies. They are later joined by two German officers, Captain Blain (Peter Sellers) and Major Schlosser (John Wayne). The Germans are ordered to take them back from the damaged village of Poitier in the southeast of France. However, they eventually discover that the British have landed, just under cover of heavy fog. Shaken by the loss of their lives, Schlosser joins in support of the French Resistance under the pretext of covering up a chemical attack on French-held Poitier. When the last Allied landing spot is discovered, Blain and Joyce use helicopter guns to guide them there. Schlosser and Joyce are placed under cover of heavy fog, but the Germans are unable to land and nearly crash-land. The escape crew, determined to stop the German air attack, attempt to land and find St. Clair. Schlosser fights his way through Poitier, but is surrounded by German soldiers. When the German forces turn towards the Germans, Blain is killed; the survivors are taken to an abandoned German village by American Private John Roberts (Richard Burton). Roberts is a spy who survived the attack; he also has a German Shepherd named Schlosser. Roberts, however, has been ordered to hunt Schlosser down. There, Roberts sees an old British pilot in a white mask who resembles Schlosser. Roberts orders Roberts to give Platoon Sergeant O'Banfield (Richard Dinklage) a position inside a Canadian-held church with two other men as hostages. The pair flee in the same boat as the others, but Roberts is captured by the Germans. When Schlosser attempts to escape, Roberts kills him. During the ensuing battle, Schlosser is killed by a machine gun, much to Joyce's disappointment. O'Banfield is forced to fight off the German forces as he is dying. Schlosser is shot and seriously wounded. After learning that Roberts has betrayed his father, Schlosser leaves Poitier to pursue Roberts, who has been killed by the German sniper. When the pair are finally able to reach Poitier, Roberts tries to stop them from moving, but Schlosser shoots Roberts and leaves him behind. During their return to Poitier, two German vehicles have crashed and the two surviving Americans are forced to swim to safety. \\
\hline
\end{longtable}

\newpage
\centering
\begin{longtable}{|m{5em}| m{13cm}|}
\caption{Generated Stories for Facts Set 3}\\  
\hline
Facts & \textbf{Fact 1:} \emph{Gomez and Morticia Addams are chased away with rest of Addams clan by angry mob} \newline
\textbf{Fact 2:} \emph{Grandmama buys Uncle Fester time to evacuate Gomez Morticia and Thing} \newline 
\textbf{Fact 3:} \emph{family refuses to change} \newline 
\textbf{Fact 4:} \emph{Pugsley struggles with complicated traditions and pressure of Mazurka even with guidance of Gomez and Uncle Fester} \newline 
\textbf{Fact 5:} \emph{Wednesday decides to run away in secret}\\
\hline
Cloze XLNet &  On 23 December 1911, British royal couple \textbf{Gomez and Morticia} (the eldest of \textbf{Addams} family) \textbf{are} successfully \textbf{chased} back \textbf{away together with the rest of the Addams} royal \textbf{clan}, presumably \textbf{by} a very \textbf{angry} Russian \textbf{mob} (the Empire (1868)). Once they reach the British, they face a number of problems: the British are worried, as they know that their number of guests will be kept in secret. \textbf{Grandmama buys Uncle Fester time to evacuate Gomez}, but he refuses to give him the money. It is left to the elderly \textbf{Morticia and Thing} to take the money. If they do, they receive no money, as the \textbf{family refuses to change} their names. They move on. When they discover that it was all a trick, they move back. After a couple of years, Gomez and Fester are both in good health, but their British leader \textbf{Puygsley} (Peter) \textbf{struggles} at home \textbf{with} the rules, \textbf{complicated} by the \textbf{traditions}, \textbf{and pressure of Mazurka}, \textbf{even with guidance of Gomez and Uncle Fester} who believe it that they need to keep the secret. Uncle \textbf{Wednesday} (Henry (Robert Owen), a British agent, \textbf{decides to run away in secret}. In desperation, he sets out with the English, where he meets a rich British girl, Dorothy (Lita Marle). She is a modern-day British woman who is in love with him, and who knows everything. She also tells him that she is in love with a young man, a former soldier. However, it appears that the young man is a revolutionary, and he also has a ``sister,'' a woman who has been sent home. The British arrive to take Mr. (Ronald) (Paul MacLane) to a far away country to give his family more time. They find that Mr. is not in Britain, and the other three are left to wait on Mr.. Meanwhile, Mr. returns. The couple move into a village, where they are welcomed.\\
\hline
BART &  Gomez and Morticia Addams are chased away with the rest of The Last Thursday by an angry mob. Gomez manages to evade arrest, but his leg is crushed when Uncle Fester (Kangaroo) tries saving him from a car crash on Highway 13 near Lake Como during which Grandmama gets hurt as well!Granddaddy explains that it was part their deal not allowing them anywhere else until he recovers completely; if they didn't return this week then she wouldn’t pay for treatment – so much money in one hand or another!! But since there were seven weeks left before her scheduled operation Wednesday decides stay put instead at home despite being called out repeatedly about leaving after five years without any explanation nor apology even though everyone lovesPugsley dearly. \\
\hline
GPT2 with pseudo-self-attention & Bella Mortimer (Deborah Kerr) and her family reside in the small town of Miffordasque, Louisiana, where she is married to the recently widowed Jim Mortimer (Alan Hale). The Mortimer family is beset with questions about why they are constantly rebuffed by their boss (Freddie Prinze). His wife, Maria Mortimer (Dorothy Quillen), is upset and urges Mortimer to seek out her grandfather. He, however, does not seem to understand, and decides to visit the family's old house, where he does not seem to understand why they are always rebuffed. The next day, Barbretasque residents are taken away as well, and the family's townspeople are forced to flee behind a wall of wood. While the police search the house for the sheriff, it is revealed that his uncle is still alive and the family has the police dead. The next day, Barbretasque is visited by one of the town boys, J. P. Ritchie (Lou Costello). Ritchie, who is wearing a bulletproof vest, is clearly upset by the presence of his uncle. Barbretasque discovers that the boys already have a date in the town's tavern, and attempts to attend the wedding. He is told by another cousin, Pizzi (John Morrellsey), that Barbretasque will have to attend an upcoming game of poker. As Barbretasque is about to enter the game, he is ambushed and his leg amputated. Barbretasque soon realizes that Ritchie caused his accident and is devastated. His last words are: ``I'd like to be with the boys.'' Barbretasque then runs down to the house. She finds her family gone. The next day, Barbretasque is gone. She discovers that the family is living in a dilapidated mansion. She attempts to visit her grandfather (Freddie Prinze), who says he can not speak to her and would rather be a doctor. Barbretasque notices Ritchie's presence and asks him to visit town, saying he is ``just in time to see the town.'' Ritchie agrees, and Barbretasque returns to the house to retrieve Ritchie's limp. Soon after, however, the townspeople begin to notice that J. P. Ritchie has been following J. P. Ritchie away. As he drives away, Barbretasque notices the ``Camelot'' in his vehicle pull over behind and collapses from the crash. When Ritchie finds out what happened, he rushes to her rescue and takes her into custody. The two are released from prison, and Barbretasque visits Jim Mortimer in the hospital. As he is returning home, she is greeted by the townspeople. After Jim says a few words about her family, Barbretasque is shot by a gunshot from behind. The townspeople are now very anxious to hear whether or not he killed her parents. They ask if his family will ever let him go, but Barbretasque refuses. \\
\hline
\end{longtable}

\end{document}